\documentclass[sigconf]{acmart}

\usepackage{acronym}
\usepackage{color}
\usepackage{amsmath}
\usepackage{amsfonts}
\usepackage[inline]{enumitem} 
\usepackage[skip=0pt]{caption}
\usepackage[ruled,vlined]{algorithm2e}
\acrodef{DS}{Dialogue System}
\acrodef{DRS}{Dialogue Response Selection}
\acrodef{DRG}{Dialogue Response Generation}
\acrodef{ODS}{Open-ended Dialogue System}
\acrodef{TDS}{Task-oriented Dialogue System}
\acrodef{KB}{Knowledge Base}
\acrodef{CF}{Collaborative Filtering}
\acrodef{NCF}{Neural Collaborative Filtering}
\acrodef{MemNN}{Memory Network}
\acrodef{SMemNN}{Split Memory Network}
\acrodef{RMemNN}{Retrieval Memory Network}
\acrodef{PMemNN}{Personalized Memory Network}
\acrodef{NPMemNN}{Neighbor-based Personalized Memory Network}
\acrodef{CoMemNN}{Cooperative Memory Network}
\acrodef{PbAbI}{personalized bAbI dialogue}
\acrodef{bAbI}{bAbI dialogue}
\acrodef{CIRS}{Conversational Information Retrieval System}
\acrodef{SOTA}{state-of-the-art}
\acrodef{MMA}{Multi-view Memory Addressing}
\acrodef{MI}{Memory Initialization}
\acrodef{MR}{Memory Reading}
\acrodef{MU}{Memory Updating}
\acrodef{DRS}{Dialogue Response Selection}
\acrodef{UPE}{User Profile Enrichment}
\acrodef{PEL}{Profile Enrichment Loss}
\acrodef{NP}{Neighbor Profile}
\acrodef{CP}{Current Profile}
\acrodef{ND}{Neighbor Dialogue}
\acrodef{CD}{Current Dialogue}
\acrodef{KNN}{K-Nearest Neighbors}
\acrodef{MLP}{Multiple Layer Perceptron}
\acrodef{RSA}{Response Selection Accuracy}
\acrodef{PEA}{Profile Enrichment Accuracy}
\newcommand{\jpei}[1]{\textcolor{black}{#1}}

\DeclareMathOperator{\ReLU}{ReLU}
\newcommand{\Softmax}{\mathrm{Softmax}}
\newcommand{\Argmax}{\mathrm{Argmax}}
\newcommand{\PiecewiseArgmax}{\mathrm{PiecewiseArgmax}}

\AtBeginDocument{%
  \providecommand\BibTeX{{%
    \normalfont B\kern-0.5em{\scshape i\kern-0.25em b}\kern-0.8em\TeX}}}

\copyrightyear{2021}
\acmYear{2021} 
\setcopyright{iw3c2w3}
\acmConference[WWW '21]{Proceedings of the Web Conference 2021}{April 19--23, 2021}{Ljubljana, Slovenia} 
\acmBooktitle{Proceedings of the Web Conference 2021 (WWW '21), April 19--23, 2021, Ljubljana, Slovenia}
\acmPrice{}
\acmDOI{10.1145/3442381.3449843}
\acmISBN{978-1-4503-8312-7/21/04}



\begin{document}

\title{A \acl{CoMemNN} for Personalized Task-oriented Dialogue Systems with Incomplete User Profiles}
%!TEX root = ./WWW21-Jiahuan-main.tex
\author{Jiahuan Pei}
\affiliation{%
  \institution{University of Amsterdam}
  \city{Amsterdam}
  \country{The Netherlands}
}
\email{j.pei@uva.nl}

\author{Pengjie Ren}
\affiliation{%
  \institution{Shandong University}
  \city{Qingdao}
  \country{China}
}
\email{renpengjie@sdu.edu.cn}
\authornote{Corresponding author}

\author{Maarten de Rijke}
\affiliation{%
  \institution{\mbox{}\hspace*{-.6cm}\mbox{University of Amsterdam \& Ahold Delhaize}}
  \city{Amsterdam}
  \country{The Netherlands}
}
\email{derijke@uva.nl}

\renewcommand{\shortauthors}{Pei et al.}

\acmSubmissionID{fp5177732}

%!TEX root = ./WWW21-Jiahuan-main.tex

\begin{abstract}
There is increasing interest in developing personalized \acp{TDS}.
Previous work on personalized \acp{TDS} often assumes that complete user profiles are available for most or even all users.
This is unrealistic because 
\begin{enumerate*}
\item not everyone is willing to expose their profiles due to privacy concerns; and
\item rich user profiles may involve a large number of attributes (e.g., gender, age, tastes, \ldots).
\end{enumerate*}
In this paper, we study personalized \acp{TDS} without assuming that user profiles are complete.
We propose a \acf{CoMemNN} that has a novel mechanism to gradually enrich user profiles as dialogues progress and to simultaneously improve response selection based on the enriched profiles.
\ac{CoMemNN} consists of two core modules: \acf{UPE} and \acf{DRS}.
The former enriches incomplete user profiles by utilizing collaborative information from neighbor users as well as current dialogues.
The latter uses the enriched profiles to update the current user query so as to encode more useful information, based on which a personalized response to a user request is selected. 

We conduct extensive experiments on the \acl{PbAbI} benchmark datasets.
We find that \ac{CoMemNN} is able to enrich user profiles effectively, which results in an improvement of 3.06\% in terms of response selection accuracy compared to state-of-the-art methods.
We also test the robustness of \ac{CoMemNN} against incompleteness of user profiles by randomly discarding attribute values from user profiles.
Even when discarding 50\% of the attribute values, \ac{CoMemNN} is able to match the performance of the best performing baseline without discarding user profiles, showing 
the robustness of \ac{CoMemNN}.
\end{abstract}

\maketitle

\acresetall

%!TEX root = ./WWW21-Jiahuan-main.tex

\section{Introduction}

The use of \acfp{TDS} is becoming increasingly widespread. 
Unlike \acp{ODS}~\citep{li2016persona,zhang2018personalizing}, \acp{TDS} are meant to help users achieve specific goals during multiple-turn dialogues~\cite{chen2017survey}.
Applications include booking restaurants, planning trips, grocery shopping, customer service~\citep[e.g.,][]{young2013pomdp,end2end_dataset_paper_babi_bordes,williams2017hybrid,madotto2018mem2seq,pei2019modular, pei2020retrospective}.

Considerable progress has been made in improving the performance of \acp{TDS}~\citep[e.g., ][]{end2end_dataset_paper_babi_bordes,liu2017gated,wu2018end,henderson2019training,  pei2019sentnet,lu2019goal,lin2020grayscale}.
Human-human dialogues reflect diverse personalized preferences in terms of, e.g., modes of expression habits~\cite{ficler2017controlling,zeng2019automatic}, individual needs and related to specific goals~\cite{bsl_memnn,mo2018personalizing,bsl_p_memnn}.
Recent work has begun to explore how to improve the user experience by  personalizing \acp{TDS} in similar ways.
Several personalized \ac{TDS} models have been proposed and have achieved good performance~\cite{bsl_memnn,bsl_p_memnn,bsl_r_memnn}.
Personalized \ac{TDS} models use user profiles in order to be able to capture, and optimize for, users' personal preferences.
Those user profiles may not always be available or complete.
While profiles may be obtained by asking users to fill in personal profiles with all predefined attributes~\cite{bsl_memnn,bsl_p_memnn,bsl_r_memnn}, more often than not, they are \textit{incomplete} and have missing values for some of the attributes of interest:
\begin{enumerate*}[topsep=0cm, leftmargin=*, itemindent=0.5cm]
\item not all users are willing to expose their profiles due to privacy concerns~\cite{wang2015collaborative}; \citet{tigunova2019listening} have shown that users rarely reveal their personal information in dialogues explicitly; and
\item user profiles may involve many attributes (such as, e.g., gender, age, tastes), which makes it hard to collect values for all of them.
For example, even if we know a user's favorite food is ``fish and chips,'' this does not mean the user does not like ``hamburgers.''
\end{enumerate*}

In this paper, we study the problem of personalized \acp{TDS} with \emph{incomplete user profiles}. 
This problem comes with two key challenges:
\begin{enumerate*}
\item how to infer missing attribute values of incomplete user profiles; and 
\item how to use enriched profiles so as to enhance personalized \acp{TDS}.
\end{enumerate*}
There have been previous attempts to extract user profiles from open-ended dialogues~\cite{tigunova2019listening,tigunova2020extracting,wu2020getting,li2020aloha,li2014personal} but to the best of our knowledge the problem of inferring and using missing attribute values has not been studied yet in the context of \acp{TDS}.

\begin{figure}[t!]
    \centering
    \includegraphics[clip,trim=0mm 0mm 12mm 0mm,width=\columnwidth]{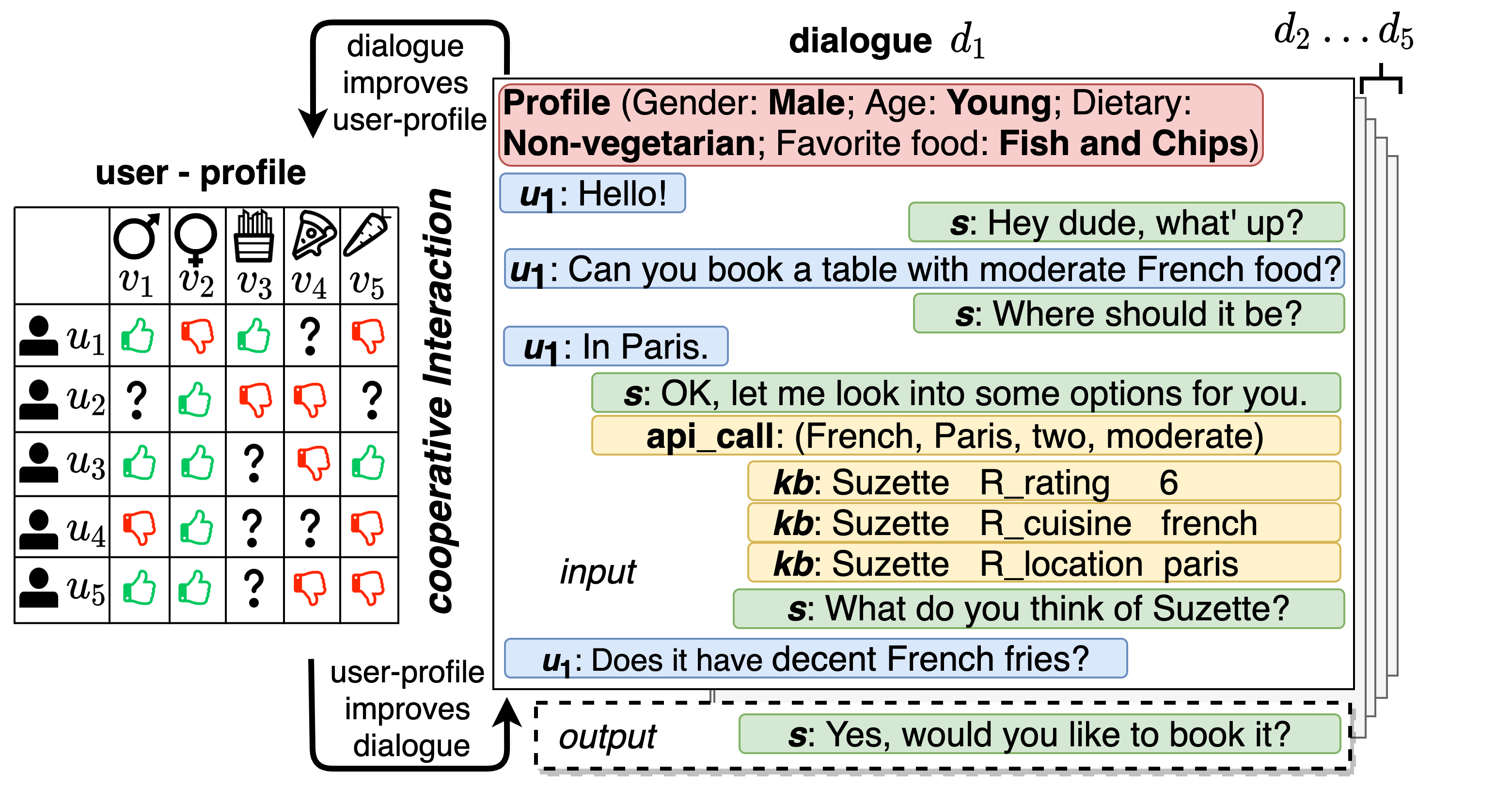}
    \caption{Cooperative interaction between user profiles and dialogues.}
    \label{fig:motivation}
\end{figure}

We address the problem of personalized \acp{TDS} with \emph{incomplete user profiles} by proposing an end-to-end \acfi{CoMemNN} in which profiles and dialogues are used to mutually improve each other. See Figure~\ref{fig:motivation} for an intuitive sketch.
The intuition behind \ac{CoMemNN} is that user profiles can be gradually improved (i.e., missing values can be added) by leveraging useful information from each dialogue turn, and, simultaneously, the performance of \ac{DRS} can be improved based on enriched profiles for later turns.
For example, when user $u_1$ produces the utterance ``Does it have `decent' french fries?'' and the user reveals his like of ``french fries,'' the attribute `favorite food' in his user profile can be enriched with the value of ``french fries.''
In addition, we want to consider collaborative information from similar users, assuming that similar users have similar preferences as reflected in their user profiles.
For example, a young male non-vegetarian who is a big fan of ``pizza'' might also love ``fish and chips'' if there are several users with similar profiles stating ``fish and chips'' as their favorite food.
In turn, knowledge of these preferences can affect the choice of the response selected by a \ac{TDS} in case there are multiple candidate responses.
In other words, users with similar profiles may expect the same or a similar response given a certain dialogue context~\citep{bsl_p_memnn}.
\ac{CoMemNN} operationalizes these intuitions with two key modules: \acf{UPE} and \acf{DRS}.
The former enriches incomplete user profiles by utilizing useful information from the current dialogue as well as collaborative information from similar users.
The latter uses the enriched profiles to update \jpei{the query representing all requested information, based on which a personalized response is selected to reply to user requests.}

To verify the effectiveness of \ac{CoMemNN}, we conduct extensive experiments on the \ac{PbAbI} benchmark dataset, which comes in two flavors, a small version \jpei{which has 1,000 dialogues}, and a large version, which has \jpei{12,000 dialogues}.
First, we find that \ac{CoMemNN} improves over the best baseline by 3.06\%/2.80\% on the small/large dataset, respectively, when using all available user profiles.
Second, to assess the performance of \ac{CoMemNN} in the presence of incomplete user profiles, we randomly discard values of attributes with varying probabilities and find that 
\jpei{even when it discards 50\% of the attribute values, the performance of \acs{CoMemNN} matches the performance of the best performing baseline without discarding user profiles.
In contrast, the best performing baseline decreases 2.12\%/1.97\% in performance on the small/large dataset with the same amount of discarded values.}

The main contributions of this paper are as follows:
\begin{itemize}[nosep,leftmargin=*]
    \item We consider the task personalized \acp{TDS} with incomplete user profiles, which has not been investigated so far, to the best of our knowledge.
    \item We devise a \acs{CoMemNN} model with dedicated modules to gradually enrich user profiles as a dialogue progresses and to improve response selection based on enriched profiles at the same time. 
    \item We carry out extensive experiments to show the robustness of \acs{CoMemNN} in the presence of incomplete user profiles.
    % \item A dataset used for aforementioned task.
\end{itemize}

%!TEX root = ./WWW21-Jiahuan-main.tex

\section{Related work}

% Personalized TDSs
In this section, we briefly present an overview of related work on personalized \acfp{ODS} and personalized \acfp{TDS}.

\subsection{Personalized \aclp{ODS}}
Previous studies on personalized \acp{ODS} mainly fuse unstructured persona information~\citep{zhang2018personalizing, mazare2018training}.
\citet{li2016persona} first attempt to incorporate a persona into the Seq2Seq framework~\cite{SEQ2SEQ} to generate personalized responses.
\citet{ficler2017controlling} apply an RNN language model conditioned on a persona to control response generation with linguistic style.
\citet{zhang2018personalizing} find that selection models based on Memory Networks~\cite{sukhbaatar2015end} are more promising than recurrent generation models based on Seq2Seq~\cite{SEQ2SEQ}.
\citet{mazare2018training} develop a response selection model based on \acs{MemNN} and model persona to improve the performance of an \acs{ODS}.
\citet{song2019exploiting} explore how to generate diverse personalized responses using a variational autoencoder conditioned on a persona memory.
\citet{liu2020you} make use of persona interaction between two interlocutors.
\citet{xu2020neural} further exploit topical information to extend persona.

Prior attempts to address data sparsity problems in order to enhance personalized \acp{ODS} have considered pretraining~\citep{herzig2017neural, zheng2020pre}, sketch generation and filling~\citep{shum2019sketch}, multiple-stage decoding~\citep{song2020generate}, multi-task learning~\citep{luan2017multi}, transfer learning~\citep{yang2017personalized,zhang2019neural,wolf2019transfertransfo}, and meta-learning~\citep{madotto2019meta}.
Only few studies have explored structured user profiles for \acsp{ODS}~\citep{qian2018assigning,zheng2019personalized,zhou2020design}.

Most of the methods listed above focus on unstructured persona information while we target structured user profiles. 
Importantly, they focus on \acp{ODS}, so they cannot be applied to \acp{TDS} directly.

\subsection{Personalized \aclp{TDS}}
Unlike \acp{ODS}, personalized \acp{TDS} have not been investigated extensively so far.
\citet{bsl_memnn} release the first and, so far, only benchmark dataset for personalized \acp{TDS}, to the best of our knowledge.
They propose a memory network based model, \acs{MemNN}, to encode user profiles and conduct personalized response selection.
They also propose an extension of \acs{MemNN}, Split \acs{MemNN}, which splits a memory into a profile memory followed by a dialogue memory.
\citet{bsl_r_memnn} introduce Retrieval \acs{MemNN} by incorporating a retrieval module into memory network, which enhances the performance by retrieving the relevant responses from other users.
\citet{bsl_p_memnn} present Personalized \acs{MemNN} which learns distributed embeddings for user profiles, dialogue history, and the dialogue history from users with the same gender and age, and shows better performance by using the idea user bias towards \ac{KB} entries over candidate responses.
\citet{mo2018personalizing} introduce a transfer reinforce learning paradigm to alleviate data scarcity, which uses a collection of multiple users as a source domain and an individual user as a target domain.

The methods above all assume that complete user profiles can be obtained by urging users to fill in all blanks in user profiles, which is unrealistic in practice. 
Thus, it remains unexplored how the methods above perform when incomplete user profiles are provided, and whether we can bridge the gap in performance if their performance is negatively affected.
An alternative is to first infer missing user profiles, e.g., by mining query logs or previous conversations~\cite{li2014personal,tigunova2019listening,tigunova2020extracting,li2020aloha}, and then apply the model with the above methods.
But to do so, we need to train a model to infer missing user profiles asynchronously.
Besides, it will likely bring cumulative errors to downstream \ac{TDS} tasks.
Instead, we propose to enrich user profiles \emph{and} achieve a \ac{TDS} simultaneously with an end-to-end model.

%!TEX root = ./WWW21-Jiahuan-main.tex

\section{Method}
\label{sec:method}

\subsection{Task}
In this work, we follow previous studies and model a personalized \ac{TDS} as a response selection task, which selects a response from predefined candidates given a dialogue context~\citep{bsl_memnn,bsl_p_memnn,bsl_r_memnn, eric2017key,wen2017network,rajendran2018learning, pei2019sentnet}. 
Table~\ref{tab:notation} summarizes the  main notation used in this paper.
\begin{table}[t]
\setlength{\tabcolsep}{1pt}
\centering
\caption{Summary of main notation used in the paper.}
\label{tab:notation}
\begin{tabular}{ll}
\toprule
$X^u_t$ & User utterance at turn $t$. \\
$X^s_t$ & System response at turn $t$. \\
$D_t$ & Dialogue history at turn $t$.\\ $\mathbf{h}_t$ & Hidden representation of $D_t$.\\
$u$ & \begin{tabular}[t]{@{}l@{}}A user profile in the form of $\{(a_i , v_i)\}_{i=1}^{m}$, $v_i$ is a candidate \\ value of $i$-{th} attribute $a_i$.\end{tabular}\\ 
$\mathbf{p}$ & One-hot representation of $u$.\\
% $k$ & Number of neighbors plus 1 for current user.\\
% $d$ & Number of embedding dimension.\\
$\mathbf{q}_t$ & \begin{tabular}[t]{@{}l@{}} A query representation that represents the user's current\\ request at turn $t$.\end{tabular}\\
$\mathbf{M}_{t}^{P}$ & \begin{tabular}[t]{@{}l@{}}Profile memory that contains user profile presentations of $u$ \\ and his/her neighbors at turn $t$. \end{tabular} \\
$\mathbf{M}_{t}^{D}$ & \begin{tabular}[t]{@{}l@{}}Dialogue memory that contains dialogue history presentation\\ of $u$ and his/her neighbors at turn $t$. \end{tabular} \\
\bottomrule
\end{tabular}
\end{table}

Given a dialogue context $(u, D_t, X^u_t)$ at the $t$-{th} dialogue turn, our goal is to select an appropriate response $y_t=X^s_{t}$ from candidate responses $Y=\{X_j^s\}|_{j=1}^{|Y|}$.
Here, $u$ is the user profile, which consists of $m$ attribute-value pairs $\{(a_i , v_i)\}_{i=1}^{m}$, where $a_i$ is the $i$-{th} attribute and $v_i$ is a candidate value of $a_i$.
For example, in Fig.~\ref{fig:motivation}, the user profile is denoted as \{(Gender, Male), (Age, Young), (Dietary, Non-vegetarian), (Favorite food, Fish and Chips)\}.
$D_t=X_{1:t-1}$ is the dialogue history.
Similar to \cite{bsl_memnn,bsl_p_memnn,bsl_r_memnn}, $D_t$ is represented as a sequence of words that are aggregated from historical utterances $[X^u_1, X^s_1, \ldots, X^u_{t-1}, X^s_{t-1}]$, alternating between the user $u$ or system $s$.
$X^u_t$ denotes the current user utterance, representing the user's current request.

\subsection{Overview of \NoCaseChange{\ac{CoMemNN}}}
\begin{figure}
    \centering
    \includegraphics[clip,trim=0mm 0mm 1mm 2mm,width=\columnwidth]{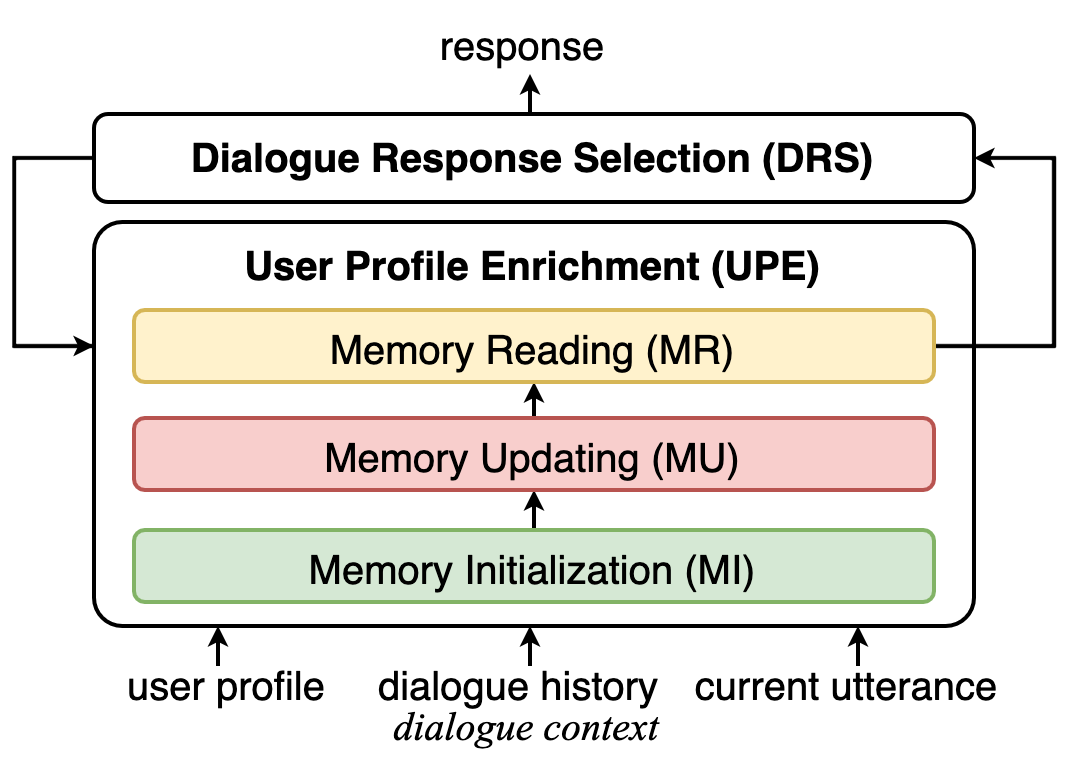}
    \caption{An overview of the \acs{CoMemNN} architecture, which consists of two cooperative modules: \acs{UPE} and \acs{DRS}.}
    \label{fig:archtecture}
\end{figure}

An overview of the proposed architecture, \ac{CoMemNN}, is shown in Fig.~\ref{fig:archtecture}.
\begin{figure*}[t!]
    \centering
    \includegraphics[clip,trim=2mm 4mm 0mm 0mm,width=2\columnwidth]{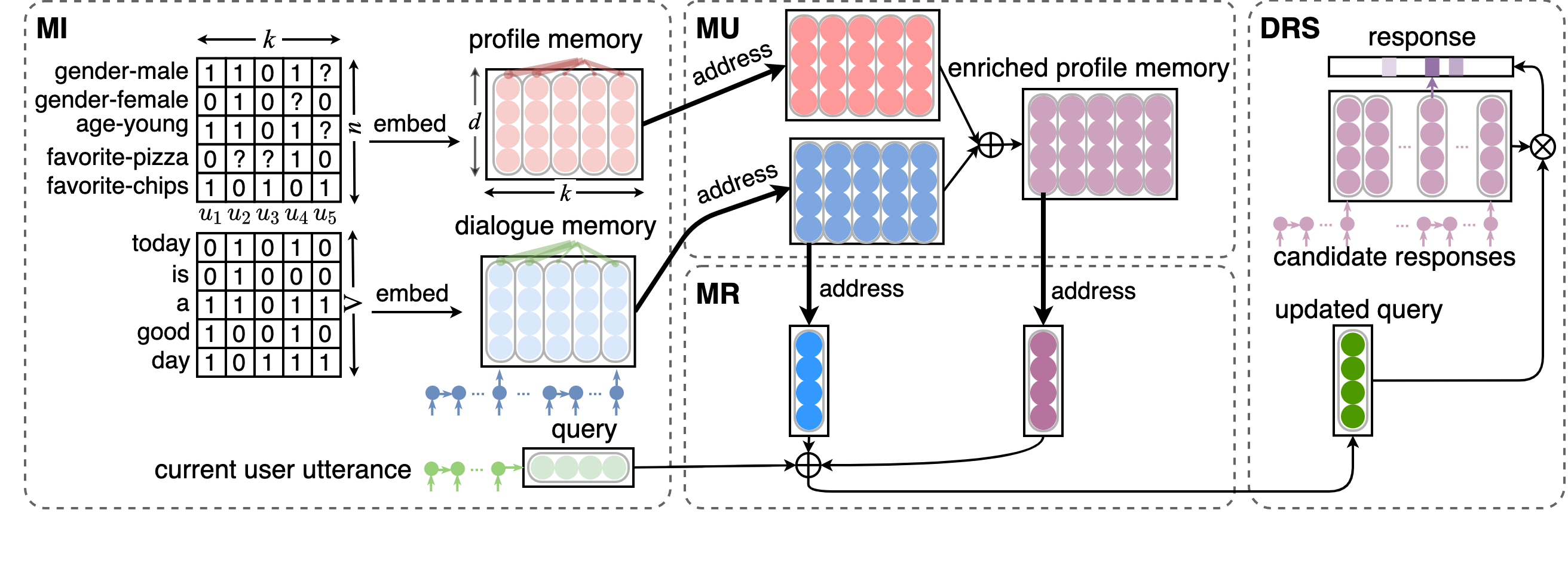}
    \caption{An overview of the dynamic pipeline of the \acs{CoMemNN} model. 
    The \acs{UPE} modules captures the interaction between user profiles and dialogues by three submodules: \acs{MI}, \acs{MU} and \acs{MR}.
    The \acs{DRS} module and the \acs{UPE} module cooperate so as to select better responses.
    Section~\ref{sec:method} contains a walkthrough of the model.}
    \label{fig:pipeline}
\end{figure*}
A key aspect of the architecture is that it aims to capture all useful information from the given dialogue context $(u, D_t, X^u_t)$, based on which we learn a query representation $\mathbf{q}_t$ to represent the user's current request.
$\mathbf{q}_t$ is usually initialized with the current user utterance $X^u_t$~\cite{bsl_memnn,bsl_p_memnn,bsl_r_memnn}.
Then, $\mathbf{q}_t$ is updated by the \acf{UPE} module by incorporating dialogue and personal information from dialogues and user profiles, respectively.
Specifically, \ac{UPE} captures the interaction between user profiles and dialogues with three submodules: \acf{MI}, \ac{MU} and \ac{MR}.
\ac{MI} searches neighbors of the current user to initialize the profile memory $\mathbf{M}_{t}^{P}$, which contains profiles from both the current user and his/her neighbors.
\ac{MI} also initializes the dialogue memory $\mathbf{M}_{t}^{D}$ with the dialogue history of both the current user and his/her neighbors, each of which is represented by addressing dialogue historical utterance representations with $\mathbf{q}_t$.
\ac{MU} updates the profile memory $\mathbf{M}_{t}^{P}$ and the dialogue memory $\mathbf{M}_{t}^{D}$ by considering their interaction, after which the user profiles are enriched by inferring missing values based on the dialogue and personal information from the current user and his/her neighbors.
Afterwards, \ac{MR} updates the query representation $\mathbf{q}_t$ by reading from the enriched profile memory as well as dialogue memory.
Finally, the \acf{DRS} module uses the updated query to match candidate responses so as to select an appropriate response.
Next, we introduce each of the modules \ac{MI}, \ac{MU} and \ac{MR}, one by one.

\subsection{\acf{MI}}
\label{subsection:MI}
\subsubsection*{Profile Memory Initialization.}
To model user-profile relations, we initialize the profile memory as: $\mathbf{M}_{t}^{P} = [\Psi(u_{1}), \dots, \Psi(u_{k})] \in \mathbb{R}^{k \times d}$, where $u_{1}$ is the \ac{CP} from the current user.
% \mdr{Do we really need those abbreivations, CP, NP? They do not appear to be used anywhere?
% \jpei{Yes, we use CP, NP, CD, ND in Section 6}
The others are \acp{NP} from neighbor users.
For each user profile, the $i$-{th} attribute can be represented as an one-hot vector $\mathbf{\tilde{p}}_i \in \mathbb{R}^{C(p_i)}$, where there are $C(a_i)$ candidate values for $p_i$.
Then, each user profile can be initialized as an one-hot vector $\mathbf{p} = \text{Concat}(\mathbf{\tilde{p}}_1, \dots, \mathbf{\tilde{p}}_m) \in \mathbb{R}^n  (n=\sum_{i=1}^{m}(C(p_i))$, which is the concatenation of one-hot representations of attributes.
$k$ is the number of users, $d$ is the embedding dimension, and $\Psi$ is a linear transformation function.
Given any user profile $u$, we find his/her $(k-1)$ nearest neighbors based on dot product similarity.

\subsubsection*{Dialogue Memory Initialization.}
\jpei{To model user-dialogue relations}, we initialize a dialogue memory $\mathbf{M}_{t}^{D} = [\mathbf{h}_{t}^{1}, \dots, \mathbf{h}_{t}^{k}] \in \mathbb{R}^{k \times d}$, where $\mathbf{h}_{t}^{1}$ is the representation of the \ac{CD} from the current user.
The others are the \acp{ND} from neighbor users.
% \mdr{Again, do we need those abbreviations?}
% \jpei{Yes.}
For each user, the dialogue history can be computed as:
\begin{equation}
    \label{eq:dmi}
    \begin{split}
        \mathbf{h}_{t} &= \sum_{i=1}^{2(t-1)} \lambda_{t}^{i} \mathbf{H}_t^{i} \in \mathbb{R}^{d}\\
        \lambda_{t}^{i} &= (\mathbf{\tilde{q}}_t)^{T} \cdot \mathbf{H}_t^{i} \in \mathbb{R}^{1},
    \end{split}
\end{equation}
where we use the updated query $\mathbf{\tilde{q}}_t$ to address the aggregated dialogue history $\mathbf{H}_t$, the addressing weight $\lambda_{t}^{i}$ is computed by the dot product of query $\mathbf{\tilde{q}}_t$ and the $i$-{th} utterance representation $\mathbf{H}_t^{i}$.

Following~\cite{dodge2015evaluating,bsl_p_memnn}, we represent each utterance as a bag-of-words using the embedding matrix $\mathbf{E} \in \mathbb{R}^{d \times V}$, where $d$ is the embedding dimension, $V$ is the vocabulary size, $\Phi(\cdot)$ maps the utterance to a bag of dimension $V$. 
At the beginning of turn $t$, the updated query $\mathbf{\tilde{q}}_t$ is initialized as:
\begin{equation}
    \mathbf{\tilde{q}}_t = \mathbf{E}\Phi(X_{t}^{u}) \in \mathbb{R}^{d}.
\end{equation}
Similarly, the aggregated dialogue history $\mathbf{H}_t$ of the current user $u_1$ can be embedded as:
\begin{equation}
\begin{split}
\mathbf{H}_t&={}\\
&[\mathbf{E}\Phi(X^u_1), \mathbf{E}\Phi(X^s_1), \dots,\mathbf{E}\Phi(X^u_{t-1}), \mathbf{E}\Phi(X^s_{t-1})] \in \mathbb{R}^{2(t-1) \times d}.
\end{split}
\end{equation}

\subsection{\acf{MU}}
\subsubsection*{Dialogue Memory Updating.}
To obtain an intermediate dialogue memory $\mathbf{\tilde{M}}_{t}^{D}$, we update the $i$-{th} dialogue memory slot $\mathbf{\tilde{M}}_{t}^{D}[:, i]$ using the newest updated query $\mathbf{\tilde{q}}_{t}$ to address initial dialogue memory $\mathbf{M}_{t}^{D}$ as:
\begin{equation}
    \label{eq:mu_dialogue}
    \begin{split}
        \mathbf{\tilde{M}}_{t}^{D}[: , i] &= \sum_{j=1}^{k} \beta_{t}^{j} \mathbf{M}_{t}^{D}[:, j] \in \mathbb{R}^{d} \\
        \beta_{t}^{j} &= (\mathbf{\tilde{q}}_{t})^{T} \cdot \mathbf{M}_{t}^{D}[:, j] \in \mathbb{R}^{1}.
    \end{split}
\end{equation}
Next, the initial dialogue memory is updated by assigning $\mathbf{M}_{t}^{D}=\mathbf{\tilde{M}}_{t}^{D}$.
As the dialogue evolves, the profile memory will gradually improve the dialogue memory because $\mathbf{\tilde{q}}_{t}$ contains information from the previous profile memory, so addressing with $\mathbf{\tilde{q}}_{t}$ \jpei{links profile-dialogue relations to the dialogue memory}.

\subsubsection*{Profile Memory Updating.}
Similarly, we can obtain an intermediate profile memory $\mathbf{\tilde{M}}_{t}^{P}$ with the following steps:
\begin{equation}
    \label{eq:mu_profile}
    \begin{split}
        \mathbf{\tilde{M}}_{t}^{P}[:, i] &= \sum_{j=1}^{k} \alpha_{t}^{j} \mathbf{M}_{t}^{P}[:, j] \in \mathbb{R}^{d} \\
        \alpha_{t}^{j} &= (\mathbf{M}_{t}^{P}[: , i])^{T} \cdot \mathbf{M}_{t}^{P}[:, j] \in \mathbb{R}^{1}.
    \end{split}
\end{equation}
Next, the profile memory slot $\mathbf{M}_{t}^{P}[:, i]$ is updated by a function $\Gamma(\cdot)$ using the intermediate profile memory slot $\mathbf{\tilde{M}}_{t}^{P}[:, i]$ and the newest updated dialogue memory slot $\mathbf{\tilde{M}}_{t}^{D}[:, i]$:
\begin{equation}
    \mathbf{M}_{t}^{P}[:, i] = \Gamma(\mathbf{\tilde{M}}_{t}^{P}[:, i], \mathbf{\tilde{M}}_{t}^{D}[:, i]) \in \mathbb{R}^{d},
\end{equation}
where $\Gamma(\cdot)$ is a mapping function that is implemented by a \ac{MLP} in this work.
In this process, the dialogue memory helps to improve the profile memory because $\Gamma(\cdot)$ \jpei{links dialogue-profile relations to the profile memory}.

\subsection{\acf{MR}}
\subsubsection*{Dialogue Memory Reading.}
Since the first memory slot corresponds to the current user, we compute $\mathbf{m}_{t}^{D}$ by hard addressing and use it to update the query $\mathbf{\tilde{q}}_{t}$ as follows:
\begin{equation}
    \label{mr_dialogue}
    \begin{split}
       \mathbf{\tilde{q}}_{t} &= \mathbf{\tilde{q}}_{t} + \mathbf{m}_{t}^{D} \in \mathbb{R}^{d} \\
       \mathbf{m}_{t}^{D} &= \mathbf{\tilde{M}}_{t}^{D}[:, 1] \in \mathbb{R}^{d}.
    \end{split}
\end{equation}

\subsubsection*{Profile Memory Reading.} Similarly, we obtain $\mathbf{m}_{t}^{P}$ by hard addressing and use it to update the query $\mathbf{\tilde{q}}_{t}$ as follows:
\begin{equation}
    \label{mr_profile}
    \begin{split}
       \mathbf{\tilde{q}}_{t} &= \mathbf{\tilde{q}}_{t} + \mathbf{m}_{t}^{P} \in \mathbb{R}^{d} \\
       \mathbf{m}_{t}^{P} &= \mathbf{M}_{t}^{P}[:, 1] \in \mathbb{R}^{d}.
    \end{split}
\end{equation}

\subsection{\acl{DRS}}
We use the latest updated query $\mathbf{\tilde{q}}_{t}$ to match with candidate dialogue responses and the predicted response distribution is computed as follows:
\begin{equation}
  \label{eq:drs}
  \begin{split}
    \mathbf{\tilde{y}}_t &= \Softmax(\mathbf{\tilde{q}}_t^T\mathbf{r}_1+\mathbf{b}_1, \dots, \mathbf{\tilde{q}}_t^T\mathbf{r}_{|Y|}+\mathbf{b}_{|Y|}) \in \mathbb{R}^{|Y|} \\
    \mathbf{b}_j &= 
        \begin{cases}
        \mathbf{f}_i \in \mathbb{R}^{1}& \text{if $\mathbf{r}_j$ mentions $i$-{th} attribute of a \acs{KB} entry} \\
        0& \text{otherwise}
        \end{cases}\\
    \mathbf{f} &=\ReLU (\mathbf{F}\mathbf{p}_1) \in \mathbb{R}^{kb},
  \end{split}
\end{equation}
where $\mathbf{r}_j$ is the representation of the $j$-{th} candidate response, $|Y|$ is the number of all candidate responses.
We follow~\citet{bsl_p_memnn} to model the user bias towards \ac{KB} entries over the $j$-{th} candidate response by a term $\mathbf{b}_j$, where the dimension 
$kb$ is the number of attributes of a \acs{KB} entry.
$\mathbf{p}_1 \in \mathbb{R}^{n}$ is the one-hot representation of the current user profile.
$\mathbf{F} \in \mathbb{R}^{kb \times n}$ maps user profiles into a \acs{KB} entry. 

\subsection{Learning of \acs{CoMemNN}}
Multiple-hop reading or updating has been shown to help improve performance of \acs{MemNN} by reading or updating the memory multiple times~\citep{sukhbaatar2015end,bsl_memnn,bsl_p_memnn}.
To enhance \ac{CoMemNN}, we devise a learning algorithm \jpei{to update the query} and memories with multiple hops, and further differentiate the specific losses of the \ac{UPE} and \ac{DRS} modules.
The learning procedure is shown in Algorithm~\ref{alg:algorithm}.
First, \ac{MI} searches neighbors $\{u_2, \dots, u_k\}$ of the current user $u_1$ to initialize the profile memory $\mathbf{M}_{t}^{P}$ and dialogue memory $\mathbf{M}_{t}^{D}$.
Second, \ac{MU} and \ac{MR} are conducted $HopN$ times, and for each time:
\ac{MU} updates the dialogue memory $\mathbf{M}_{t}^{D}$ and the profile memory $\mathbf{M}_{t}^{P}$ by considering their cooperative interaction.
After that, \ac{MR} updates the query representation $\mathbf{q}_t$ by reading from the enriched dialogue memory followed by profile memory.
Last, the \acf{DRS} module uses the newest updated query $\mathbf{\tilde{q}}_t$ to match candidate responses so as to predict a response distribution $\mathbf{\tilde{y}}_t$.

To evaluate the performance of \acs{DRS} and \acs{UPE}, we define two mapping functions to get prediction labels:
\begin{itemize}[leftmargin=*,nosep]
    \item $\Argmax (\cdot)$: it outputs the index $y_t$ with the highest probability in a predicted response distribution $\tilde{\mathbf{y}}_t$;
    \item $\PiecewiseArgmax (\cdot)$: it generates a 1-0 vector from the predicted enriched profile $\mathbf{m}_{t}^{P}$, where $\tilde{\mathbf{p}}_t^1[i]=1$ only if $\mathbf{m}_{t}^{P}[i]$ achieves the highest probability among the values that belong to the same attribute.
\end{itemize}
To optimize \acs{DRS}, we use a standard cross-entropy loss between the prediction $\mathbf{\tilde{y}}$ and the one-hot encoded true label $\mathbf{y}$:
\begin{equation}\label{eq:drs_loss}
    \mathcal{L}_{\acs{DRS}} (\theta) = - \frac{1}{N_1} \sum_{i=1}^{N_1} \sum_{j=1}^{|Y|} \mathbf{y}_j \log \mathbf{\tilde{y}}_j,
\end{equation}
where $\theta$ are all parameters in the model and $N_1$ is the number of training samples.

To control the learning of \ac{UPE}, we introduce the element-wise mean squared loss between the sampled profile $\mathbf{p} = \{p_1, \dots, p_{N_2}\}$ and its corresponding enriched profile $\tilde{\mathbf{p}} = \{\tilde{p}_1, \dots, \tilde{p}_{N_2}\}$:
\begin{equation}\label{eq:upe_loss}
    \mathcal{L}_{\acs{UPE}} (\theta) = - \frac{1}{N_2}\sum_{i=1}^{N_2}(p_i - \tilde{p}_i),
\end{equation}
where $\theta$ are all parameters in the model and $N_2$ is the number of sampled values.

Finally, the final loss is a linear combination:
\begin{equation}\label{eq:final_loss}
    \mathcal{L}(\theta) = \mu \mathcal{L}_{\acs{DRS}}(\theta) + (1-\mu) \mathcal{L}_{\acs{UPE}}(\theta),
\end{equation}
where $\mu$ is a hyper-parameter to balance the relative importance of the constituent losses.

\begin{algorithm}[t!]
\caption{Multiple hop \acs{CoMemNN}.}
\label{alg:algorithm}
\SetAlgoLined
\LinesNumbered 
\SetKwComment{Comment}{$\triangleright$\ }{}
\KwIn{
turn $t$, 
user $u_1$, 
profile $\mathbf{p}_1$, 
dialogue history $\mathbf{H}_t$, 
query $\mathbf{q}_t$, 
response candidates $\{\mathbf{r}_{1}, \dots, \mathbf{r}_{|Y|} \}$,
max hop $HopN$,
$(k-1)$ neighbors
}
\KwOut{A index $\mathbf{y}_t$ of next response; An one-hot vector $\tilde{\mathbf{p}}_{t}^{1}$ presenting the enriched profile.}
$\{u_2, \dots, u_k\} \leftarrow \text{Search}(\mathbf{p}_1, k-1)$\Comment*[r]{\acs{MI}}
$\mathbf{M}_{t}^{P} \leftarrow [\mathbf{p}_{1}, \dots, \mathbf{p}_{k}] $\; 
% $\mathbf{\tilde{q}}_{t} \leftarrow \mathbf{q}_t$\;
$\mathbf{M}_{t}^{D} \leftarrow [\mathbf{h}_{t}^{1}, \dots, \mathbf{h}_{t}^{k}]; \mathbf{h}_{t}^{i} \leftarrow (\tilde{q}_{t}, \mathbf{H}_{t}^{i}), i \in [1, k];
\mathbf{\tilde{q}}_{t} \leftarrow \mathbf{q}_t$\; 
\While{hop $\leq$ HopN}{
$\mathbf{\tilde{M}}_{t}^{D} \leftarrow \mathbf{M}_{t}^{D}$; $\mathbf{\tilde{M}}_{t}^{P} \leftarrow \mathbf{M}_{t}^{P}$ \Comment*[r]{\acs{MU}}
$\mathbf{M}_{t}^{D} \leftarrow \mathbf{\tilde{M}}_{t}^{D}$\;
$\mathbf{M}_{t}^{P} \leftarrow \Gamma( \mathbf{\tilde{M}}_{t}^{P}, \mathbf{\tilde{M}}_{t}^{D})$\;
$\mathbf{m}_{t}^{D} \leftarrow \mathbf{M}_{t}^{D}$; $\mathbf{\tilde{q}}_{t} \leftarrow \mathbf{\tilde{q}}_{t} + \mathbf{m}_{t}^{D}$ \Comment*[r]{\acs{MR}}
$\mathbf{m}_{t}^{P} \leftarrow \mathbf{M}_{t}^{P}$; $\mathbf{\tilde{q}} \leftarrow \mathbf{\tilde{q}}_{t} + \mathbf{m}_{t}^{P}$\;
}
$\mathbf{\tilde{y}}_t \leftarrow \Softmax(\mathbf{\tilde{q}}_t^T\mathbf{r}_1+\mathbf{b}_1, \dots, \mathbf{\tilde{q}}_t^T\mathbf{r}_{|Y|}+\mathbf{b}_{|Y|})$ \Comment*[r]{DRS}
$y_t \leftarrow \Argmax_j (\mathbf{\tilde{y}}_t)$\;
$\tilde{\mathbf{p}}_{t}^{1} \leftarrow \PiecewiseArgmax (\mathbf{m}_{t}^{P})$
\end{algorithm}

%!TEX root = ./WWW21-Jiahuan-main.tex

\section{Experimental Setup}
\subsection{Research questions}
We seek to answer the following questions in our experiments:
\begin{enumerate*}[label=(Q\arabic*)]
\item How well does \acs{CoMemNN} perform? Does it significantly and continuously outperform state-of-the-art methods? 
\item What are the effects of different components in \acs{CoMemNN}?
\item Do different profile attributes contribute differently? and
\item \jpei{How well does \acs{CoMemNN} perform in terms of robustness?}
\end{enumerate*}

\subsection{Dataset and evaluation}
We use the \acf{PbAbI} dataset~\cite{bsl_memnn} for our experiments; this is an extension of the \ac{bAbI} dataset that incorporates personalization~\cite{end2end_dataset_paper_babi_bordes}.
To the best of our knowledge, this is the only available open dataset for personalized \acp{TDS}.
There are two versions: a large version with around 12,000 dialogues and a small version with 1,000 dialogues.
These two datasets share the same vocabulary with $14,819$ tokens and candidate response set with $43,863$ responses.
It defines four user profile attributes (gender, age, dietary preference, and favorite food) and composes corresponding attribute-value pairs to a user profile.
Each conversation is provided with all of the above user profile attributes, e.g., \{(Gender, Male), (Age, Young), (Dietary, Non-vegetarian), (Favorite: Fish and Chips)\}.
But this does not mean the given user profile is complete because the user may also like ``Paella'', although ``Fish and Chips'' is his/her favorite food.
To simulate incomplete profiles with various degrees of incompleteness, we randomly discard attribute values from a user profile with probabilities of [0\%, 10\%, 30\%, 50\%, 70\%, 90\%, 100\%] and obtain 7 alternative datasets, respectively.

We evaluate the performance of the full dialogue task using the following two metrics~\citep{bsl_memnn}:
\begin{itemize}[leftmargin=*,nosep]
    \item \acfi{RSA}: the fraction of correct responses out of all candidate responses~\citep{bsl_memnn,bsl_p_memnn}; and
    \item \acfi{PEA}: we define this metric as the fraction of correct profile values out of all discarded profile values.
\end{itemize}
We use a paired t-test to measure statistical significance ($p <0.01$) of relative improvements.

To compare model stability, we propose a statistic $\sigma$, namely \textit{stability coefficient}, which is defined as the standard deviation of a list of performance results.
Formally, given a list of evaluation values $[z_1, \dots, z_{N+1}]$, either \acs{RSA} or \acs{PEA} scores, $\sigma$ is computed as follows:
\begin{equation}
    \label{eq:sigma}
    \begin{split}
        \sigma(\mathbf{z}) &= \sqrt{\frac{1}{N}\sum_{i=1}^{N}{(\mathbf{z}_{i}-
\bar{\mathbf{z}})^2}} \\
        \mathbf{z} &= [z_2-z_1, \dots, z_{N+1}-z_{N}],
    \end{split}
\end{equation}
where $\bar{z}$ is the mean of the values in performance difference list $\mathbf{z}$.

\subsection{Baselines}
We compare with all the methods that have reported results on the \ac{PbAbI} dataset~\cite{bsl_memnn}.
\begin{itemize}[leftmargin=*,nosep]
\item \textbf{\ac{MemNN}}. 
    It regards the profile information as the first user utterance ahead of each dialogue and achieves personalization by modeling dialogue context using the standard \acs{MemNN} model~\cite{bordes2016learning}. 
\item \textbf{\ac{SMemNN}}. 
    It splits memory into a proﬁle memory and a dialogue memory. 
    The former encodes user profile attributes as separate entries and the latter operates the same as the \acs{MemNN}.
    The element-wise sum of both memories are used for ﬁnal decision~\cite{bsl_memnn}.
\item \textbf{\ac{RMemNN}}. 
    It features an en\-coder-encoder memory network with a retrieval module that employs the user utterances and user profiles to collect relevant information from similar users' conversations~\cite{bsl_r_memnn}.
\item \textbf{\ac{PMemNN}}.
    It uses \acs{MemNN} to model the current user profile, the current dialogue history, as well as the dialogue history of all users with the same gender and age.
    It also models user bias towards different \ac{KB} entries~\cite{bsl_p_memnn}.
\item \textbf{\acl{NPMemNN}} \textbf{(NP\-Mem\-NN)}. 
    Our implementation of \acs{PMemNN} is based on Pytorch.
    Unlike \acs{PMemNN}, we use the dialogue history from the nearest $(k-1)$ neighbors instead of all users with the same gender and age.
\end{itemize}

\subsection{Implementation details}
We follow the experimental settings detailed in~\citep{bsl_p_memnn}.
The embedding size of word/profile is 128.
The size of memory is 250.
The mini-batch size is 64.
The maximum number of training epoch is 250, and the number of hops is 3 (see Algorithm~\ref{alg:algorithm}).
The \ac{KNN} algorithm is implemented based on faiss\footnote{\url{https://github.com/facebookresearch/faiss}} with the inner product measurement and the number of collaborative users $k=100$.
We implement \acs{NPMemNN} and \ac{CoMemNN} in PyTorch.\footnote{\url{https://pytorch.org/}}
And the code of the other models is taken from the original papers.
We use Adam~\cite{adam_optimizer} as our optimization algorithm with learning rate of $0.01$ and initialize the learnable parameters with the Xavier initializer.
We also apply gradient clipping \cite{pmlr-v28-pascanu13} with range $[-10, 10]$ during training.
We use $l2$ regularization to alleviate overfitting, the weight of which is set to $10^{-5}$.
We treat the importance of losses of \acs{DRS} and \acs{UPE} equally, i.e., $\mu=0.5$.
The code is available online.\footnote{\url{https://github.com/Jiahuan-Pei/CoMemNN}}

%!TEX root = ./WWW21-Jiahuan-main.tex

\section{Results (Q1)}

\subsection{Results without discarding user profiles}

We show the overall response selection performance of all methods in Table~\ref{tab:main_comparison_with_baselines}. 
\begin{table}[h]
\centering
\caption{Overall performance in terms of the \acs{RSA} metric. \textbf{Bold face} indicates leading results. Significant improvements over \acs{NPMemNN} are marked with $^\ast$ (paired t-test, $p < 0.01$).}
\label{tab:main_comparison_with_baselines}
\begin{tabular}{lcc}
\toprule
                    & \multicolumn{1}{l}{\bf Small set (\%)} & \multicolumn{1}{l}{\bf Large set (\%)} \\ \midrule
\acs{MemNN}~\cite{bsl_memnn}              & 77.74                         & 85.10                        \\
\acs{SMemNN}~\cite{bsl_memnn}       & 78.10                         & 87.28                        \\
\acs{RMemNN}~\cite{bsl_r_memnn}    & 83.94                         & 87.33                        \\
\acs{PMemNN}~\cite{bsl_p_memnn} & 88.07                         & 95.33                        \\ \midrule
\acs{NPMemNN}       & 87.91                         & 97.49                        \\
\acs{CoMemNN}           & \textbf{91.13}\rlap{$^\ast$}                         & \textbf{98.13}\rlap{$^\ast$}                       \\ \bottomrule
\end{tabular}
\end{table}

\noindent%
First, \acs{CoMemNN} outperforms all baselines on both the small and large datasets by a large margin.
It significantly outperforms the best baseline \ac{PMemNN} by 3.06\% on the small dataset and 2.80\% on the large dataset.
The improvements demonstrate the effectiveness of \acs{CoMemNN}.
We believe the main reason is that the proposed cooperative mechanism is able to enrich the incomplete profiles gradually as dialogues progress and the enriched profiles improve help to response selection simultaneously.
We will analyze this in more depth in the next session.

Second, the performance of \acs{NPMemNN} is comparable to that of \acs{PMemNN} on the small dataset and achieves 2.16\% higher \acs{RSA} on the large dataset.
Recall that \acs{NPMemNN} is our implementation of \acs{PMemNN} using Pytorch; the only difference is the KNN algorithm used for neighbor searching, so the result shows that our new neighbor searching method is more effective.
Since our \acs{CoMemNN} is built upon \acs{NPMemNN}, for the remaining experiments, we will use \acs{NPMemNN} for further comparison and analysis.

Third, the results on the small and large datasets mostly show consistent trends.
For the remaining analysis experiments in the next section (Section~\ref{sec:analysis}), we will report results on the small dataset only. The findings on the large dataset are qualitatively similar.

\subsection{Results with different profile discard ratios}

We compare \acs{CoMemNN} and \acs{NPMemNN} under different profile discard ratios.
The results are shown in Table~\ref{tab:main_comparison_discard_ratios}. 
\begin{table}[h]
\setlength{\tabcolsep}{3pt}
\centering
\caption{Comparison of \acs{CoMemNN} and \acs{NPMemNN} in terms of the \ac{RSA} metric w.r.t.\ different profile discard ratios. \textbf{Bold face} indicates leading results. Significant improvements over \acs{NPMemNN} are marked with $^\ast$ (paired t-test, $p < 0.01$). 
The values of Diff. are computed by absolute difference of \acs{RSA} (\%) between \acs{CoMemNN} and \acs{NPMemNN}.}
\label{tab:main_comparison_discard_ratios}
\begin{tabular}{@{}lccccccc@{}}
\toprule
\textbf{Discard Ratio} & 0\%   & 10\%  & 30\%  & 50\%  & 70\%  & 90\%  & 100\% \\ \midrule
\acs{NPMemNN}   & 87.91 & 86.11 & 86.56 & 85.79 & 83.93 & 84.08 & \textbf{84.83} \\
\acs{CoMemNN}    & \textbf{91.13}\rlap{$^\ast$}  & \textbf{89.90}\rlap{$^\ast$} & \textbf{88.69}\rlap{$^\ast$}  & \textbf{87.80}\rlap{$^\ast$}  & \textbf{86.35}\rlap{$^\ast$}  & \textbf{84.83}\rlap{$^\ast$}  & 82.85 \\ \midrule
Small Set/Diff. & \phantom{0}3.22 & \phantom{0}3.79 & \phantom{0}2.13 & \phantom{0}2.01 & \phantom{0}2.42 & \phantom{0}0.75 & $-$1.98\\ 
\bottomrule
\acs{NPMemNN}    & 97.49 & 97.01 & 96.05 & 95.52 & 95.40 & 90.96 & 90.50 \\
\acs{CoMemNN}     & \textbf{98.13}\rlap{$^\ast$} & \textbf{97.94}\rlap{$^\ast$} & \textbf{97.68}\rlap{$^\ast$} & \textbf{97.53}\rlap{$^\ast$} & \textbf{96.98}\rlap{$^\ast$} & \textbf{96.63}\rlap{$^\ast$} & \textbf{92.73}\rlap{$^\ast$} \\
\midrule
Large Set/Diff. & \phantom{0}0.64 & \phantom{0}0.93 & \phantom{0}1.63 & \phantom{0}2.01 & \phantom{0}1.58 & \phantom{0}5.67 & \phantom{0}2.23
\\
\bottomrule
\end{tabular}
\end{table}

First, \acs{CoMemNN} significantly outperforms \acs{NPMemNN} on both the small and large datasets when the profile discard ratios range from 0\% to 90\%.
Specifically, it gains an improvement of 0.75\%--3.79\% on the small dataset and 0.64\%--5.67\% on the large dataset, respectively.
Without discarding profile attribute values, \acs{CoMemNN} achieves 3.22\% / 0.64\% of improvement compared with \acs{NPMemNN}.
Unlike the raw profiles where each attribute has only one value, the enriched profile generated by \acs{CoMemNN} is able to represent a distribution over all possible values, which can better capture users' preference.
For example, a user may label ``Fish and Chips'' as his favorite food, but this does not mean he does not like ``Paella.''
With the raw profile, this is not addressed.

Second, the performance of \acs{CoMemNN} steadily decreases with the increase of the profile discard ratio, as is to be expected.
This is reasonable as it becomes more and more challenging for \acs{CoMemNN} to find back missing values of user profiles.
Interestingly, the performance difference between \acs{CoMemNN} and \acs{NPMemNN} first increases and then decreases with the increase of the profile discard ratio.
A possible reason is that \acs{CoMemNN} is able to infer the missing values of user profiles effectively with lower profile discard ratios.
However, the profile enrichment ability decreases due to the lack of too many profile values. 
This hypothesis can be verified by the results that the increase trend lasts longer on the large dataset.
Because even with the same profile discard ratio, there are more values of user profiles left on the large dataset for \acs{CoMemNN} to infer the missing ones.
We note that \acs{NPMemNN} outperforms \acs{CoMemNN} when all user profiles are discarded on the small dataset.
The reason is that \ac{UPE} cannot enrich user profiles properly in this case, which results in a negative impact on \acs{DRS}.
But this is not the case on the large dataset where \ac{UPE} can still enrich user profiles properly when the model can find enough personal information clues from more dialogue history.

Third, to answer \textbf{Q4}, we compute the statistic $\sigma$ (Eq.~\ref{eq:sigma}) to compare the model stability.
The $\sigma$ values for \acs{CoMemNN} and \acs{NPMemNN} are 0.3357/1.0407 on the small dataset and 1.3479/1.4849 on the large dataset, respectively.
Thus, \acs{NPMemNN} has higher deviations, which shows that \acs{CoMemNN} is more stable than \acs{NPMemNN} with various profile discard ratios. 

%!TEX root = ./WWW21-Jiahuan-main.tex

\section{Analysis}
\label{sec:analysis}

We analyze the performance of the following variants of \acs{CoMemNN}:
\begin{itemize}[leftmargin=*,nosep]
\item \textbf{\acs{CoMemNN}}. The full model.
\item \textbf{\acs{CoMemNN}-\acs{PEL}}. \acs{CoMemNN} without \acf{PEL}, defined in Eq.~\ref{eq:upe_loss}.
\item \textbf{\acs{CoMemNN}-\acs{PEL}-\acs{UPE}}. \acs{CoMemNN} without \ac{PEL} or \acs{UPE}. This is exactly \acs{NPMemNN}.
\item \textbf{\acs{CoMemNN}-\acs{NP}}. \acs{CoMemNN} without the \acf{NP} as  input for \acs{UPE}.
\item \textbf{\acs{CoMemNN}-\acs{NP}-\acs{CP}}. \acs{CoMemNN} without \ac{NP} or the \acf{CP} as input for \acs{UPE}.
\item \textbf{\acs{CoMemNN}-\acs{ND}}. \acs{CoMemNN} without the \acf{ND} of dialogues as input for \acs{UPE}.
\item  \textbf{\acs{CoMemNN}-\acs{ND}-\acs{CD}}. \acs{CoMemNN} without \acs{ND} or the \acf{CD} of dialogues as input for \acs{UPE}. 
\item  \textbf{\acs{CoMemNN}-\acs{ND}-\acs{NP}}. \acs{CoMemNN} without \acs{ND} or \acs{NP} of dialogues as input for \acs{UPE}. 
\end{itemize}

\subsection{Ablation study on \acs{PEA} (Q2)}

We study the \ac{PEA} performance of different variants in Table~\ref{tab:analysis_ablation_study_UPE}. 
\begin{table*}[htb!]
\setlength{\tabcolsep}{12.5pt}
\centering
\caption{Performance of \ac{UPE} evaluated in terms of \acf{PEA}. In each cell, the ﬁrst number represents the \acs{PEA} (\%), and the number in parentheses shows the difference compared with \acs{CoMemNN}. 
$\downarrow$ and $||$ denote a decrease and no change compared to \acs{CoMemNN}, respectively. }
\label{tab:analysis_ablation_study_UPE}
\begin{tabular}{@{}lcccccc@{}}
\toprule
\textbf{Discard Ratio}   & 10\%  & 30\%  & 50\%  & 70\%  & 90\%  & 100\% \\ \midrule
\ac{CoMemNN} & 99.99 & 99.93 & 99.82 & 99.83 & 99.38 & 98.98 \\ \midrule
\acs{CoMemNN}-\acs{PEL} & 85.71 ($\downarrow$14.28) & 87.85 ($\downarrow$12.08) & 91.34 ($\downarrow$8.48) & 89.19 ($\downarrow$10.64) & 90.04 ($\downarrow$9.34)                        & 90.60 ($\downarrow$8.38) \\ \midrule
\acs{CoMemNN}-\acs{NP}                                                                            & 99.87 ($\downarrow$0.12) & 99.85 ($\downarrow$0.08) & 99.24 ($\downarrow$0.58) & 99.15 ($\downarrow$0.68)                        & 99.13 ($\downarrow$0.25)                        & 98.86 ($\downarrow$0.12) \\
\acs{CoMemNN}-\acs{NP}-\acs{CP} & 98.89 ($\downarrow$1.10) & 99.09 ($\downarrow$0.84) & 99.16 ($\downarrow$0.66) & 99.20 ($\downarrow$0.63)                        & 99.14 ($\downarrow$0.23)                        & 98.92 ($\downarrow$0.06) \\
\acs{CoMemNN}-\acs{ND} & 99.72 ($\downarrow$0.26) & 99.87 ($\downarrow$0.06) & 99.80 ($\downarrow$0.02) & 99.46 ($\downarrow$0.37) & 98.72 ($\downarrow$0.66) & 97.23 ($\downarrow$1.75) \\
\acs{CoMemNN}-\acs{ND}-\acs{CD} & 99.99 ($||$0.00) & 99.86 ($\downarrow$0.07) & 99.68 ($\downarrow$0.14) & 99.69 ($\downarrow$0.14)                        & 99.19 ($\downarrow$0.19)                        & 34.78 ($\downarrow$64.2) \\ 
\acs{CoMemNN}-\acs{ND}-\acs{NP} & 99.09 ($\downarrow$0.90) & 98.98 ($\downarrow$0.95) & 97.95 ($\downarrow$1.87) & 97.69 ($\downarrow$2.14)                        & 97.06 ($\downarrow$2.32)                        & 97.23 ($\downarrow$1.75) \\ 
\bottomrule
\end{tabular}
\end{table*}
\begin{table*}[htb!]
\setlength{\tabcolsep}{8pt}
\centering
\caption{Ablation study on \acs{DRS} evaluated in terms \acf{RSA}. In each cell, the ﬁrst number represents the \acs{RSA} (\%), and the number in parentheses shows the difference compared with \acs{CoMemNN}. 
$\downarrow$ and $\uparrow$ denote decrease and increase, respectively. Underlining marks results that are $\geq$1.0\% higher than those of \acs{CoMemNN}.}
\label{tab:analysis_ablation_study_DRS}
\begin{tabular}{@{}lccccccc@{}}
\toprule
\textbf{Discard Ratio}                             & 0\%   & 10\%  & 30\%  & 50\%  & 70\%  & 90\%  & 100\% \\ \midrule
\ac{CoMemNN}                           & 91.13 & 89.90 & 88.69 & 87.80 & 86.35 & 84.83 & 82.85 \\ \midrule
\acs{CoMemNN}-\acs{PEL}              & 90.84 ($\downarrow$0.29) & 90.29 ($\uparrow$0.39) & 
89.07 ($\uparrow$0.38) & 87.18 ($\downarrow$0.62) & 85.42 ($\downarrow$0.93) & 80.54 ($\downarrow$4.29) & 81.23 ($\downarrow$1.62) \\ 
\acs{CoMemNN}-\acs{PEL}-\acs{UPE}  & 87.91 ($\downarrow$3.22) & 86.11 ($\downarrow$3.79) & 86.56 ($\downarrow$2.13) & 85.79 ($\downarrow$2.01) & 83.93 ($\downarrow$2.42) & 84.08 ($\downarrow$0.75) & \underline{84.83}  ($\uparrow$1.98) \\ \midrule
\acs{CoMemNN}-\acs{NP}               & 91.06 ($\downarrow$0.07) & \underline{91.23} ($\uparrow$1.33) & 89.17 ($\uparrow$0.48) & 85.26 ($\downarrow$2.54) & 83.30 ($\downarrow$3.05) & 82.10 ($\downarrow$2.73) & 82.83 ($\downarrow$0.02) \\
\acs{CoMemNN}-\acs{NP}-\acs{CP}    & 86.60 ($\downarrow$4.53) & 86.10 ($\downarrow$3.80) & 84.56 ($\downarrow$4.13) & 83.53 ($\downarrow$4.27) & 82.48 ($\downarrow$3.87) & 81.95 ($\downarrow$2.88) & 81.35 ($\downarrow$1.50) \\
\acs{CoMemNN}-\acs{ND}              & 90.91 ($\downarrow$0.22) & 87.33 ($\downarrow$2.57) & 89.06 ($\uparrow$0.37) & 87.49 ($\downarrow$0.31) & 86.59 ($\uparrow$0.24) & 85.38 ($\uparrow$0.55) & \underline{85.41} ($\uparrow$2.56) \\
\acs{CoMemNN}-\acs{ND}-\acs{CD}  & 87.70 ($\downarrow$3.43) & 90.44 ($\uparrow$0.54) & 85.79 ($\downarrow$2.90) & 84.90 ($\downarrow$2.90) & 83.56 ($\downarrow$2.79) & 82.57 ($\downarrow$2.26) & \underline{85.38} ($\uparrow$2.53) \\ 
\acs{CoMemNN}-\acs{ND}-\acs{NP}  & 90.04 ($\downarrow$1.09) & 91.08 ($\uparrow$1.18) & 89.23 ($\uparrow$0.54) & 87.38 ($\downarrow$0.42) & 85.76 ($\downarrow$0.59) & 85.46 ($\uparrow$0.63) & \underline{85.41} ($\uparrow$2.56) \\ 
\bottomrule
\end{tabular}
\end{table*}

First, \acs{CoMemNN} can effectively enrich user profiles by inferring the missing values.
It is able to correctly predict more than 98.98\% of missing values in user profiles under different profile discard ratios.
We believe \ac{UPE} benefits a lot from modeling the interaction between user profiles and dialogues.
\ac{UPE} is able to capture more personal information from dialogue history with dialogues gradually going on.
The \ac{PEA} scores are all very high, because the \ac{PbAbI} dataset is simulated, which makes it relatively easy to predict missing attribute values of user profiles.

Second, we can see that each component of \acs{UPE} generally has a positive effect on the performance since most \acs{PEA} scores of most variants decrease.
Specifically, \acs{CoMemNN}-\acs{PEL} decreases by 8.38\%--14.20\% compared with \acs{CoMemNN}.
This means that it is important to add the \acs{UPE} loss (Eq.~\ref{eq:upe_loss}), rather than only optimizing the \acs{DRS} loss (Eq.~\ref{eq:drs_loss}).
We also show how the four components of \acs{UPE} (i.e., \acs{NP}, \acs{CP}, \acs{ND}, and \acs{CD} as defined in Section~\ref{subsection:MI}) affect its performance.
We find that: 
\begin{enumerate*}
\item \acs{CoMemNN}-\acs{ND}-\acs{NP} continuously decreases 0.90\%--2.32\% with the increase of the profile discard ratio.
This means that neighbor users play an important role.
\item \acs{CoMemNN}-\acs{ND}-\acs{CD} (with 100\% profile discard ratio) decreases dramatically, which is as expected, because \acs{CoMemNN} cannot infer the missing values without any dialogue history and profiles. 
This also explains the increase of the corresponding \ac{RSA} score in Table~\ref{tab:analysis_ablation_study_DRS}.
\item The decrease is mostly less than 2.32\% except that the decrease of \acs{CoMemNN}-\acs{ND}-\acs{CD} (with 100\% profile discard ratio, i.e., no \acs{NP} or \acs{CP} as well) is 64.2\%.
This reveals that different information sources are complementary to each other.
The performance will not be affected largely unless all the four inputs  (i.e., \acs{NP}, \acs{CP}, \acs{ND}, \acs{CD}) are removed.
\end{enumerate*}

Lastly, we compute the stability coefficient $\sigma$ (Eq.~\ref{eq:sigma}) of the variants in Table~\ref{tab:analysis_ablation_study_UPE} which are 0.1867, 1.8781, 0.2236, 0.1402, 25.6845, 0.1867, 0.4182, respectively. 
This shows that all variants are robust in terms of the performance of \acs{UPE} with small stability coefficient, except for \acs{CoMemNN}-\acs{ND}-\acs{CD}.

\subsection{Ablation study on \ac{RSA} (Q2)}
We investigate the \ac{RSA} performance of different variants in Table~\ref{tab:analysis_ablation_study_DRS}.%

First, the performance decreases generally by removing any component of \acs{UPE}.
In particular, \acs{CoMemNN}-\acs{PEL} has a greater effect on \acs{RSA} when the profile discard ratios get larger.
This is reasonable because the larger the profile discard ratio, the more space for improvement the proposed model has compared with \acs{NPMemNN}.
\acs{CoMemNN}-\acs{PEL}-\acs{UPE} is inferior to \acs{CoMemNN}-\acs{PEL} generally, which means that the \acs{UPE} module helps as it implicitly impact the \acs{DRS} loss (Eq.~\ref{eq:drs_loss}).
But this ability weakens when the profile discard ratio is larger than 90\%.

Second, we observe that the four information sources (i.e., \acs{NP}, \acs{CP}, \acs{ND}, \acs{CD}) have different effects under different profile discard ratios.
Particularly, the profiles of the current users and their neighbors generally contribute most to the \acs{RSA} performance.
We can see that \acs{CoMemNN}-\acs{NP}-\acs{CP} drops 1.50\%--4.53\% under all profile discard ratios.
The reason is that user profiles directly store personal information; it is easier to infer missing values from collaborative user profiles than from dialogues.

Third, we find that \acs{NP} and \acs{ND} are complementary to each other. 
\acs{CoMemNN}-\acs{NP} either has a massive drop (2.54\%--3.05\%) or small changes ($\leq$0.48\%) with the most profile discard ratios, except for one obvious rise (1.33\%) under the 10\% profile discard ratio.
In contrast, \acs{CoMemNN}-\acs{ND} works fine under the 10\% profile discard ratio, but it performs poorly for the rest.
Thus, the performance of \acs{CoMemNN} is influence strongly by a drop in attribute values unless we remove both \acs{NP} and \acs{ND} under 100\% profile discard ratios.

Lastly, the dialogue history also contributes to the \acs{RSA} performance in most cases.
\acs{CoMemNN}-\acs{ND}-\acs{CD} shows decrease (2.26\%--3.43\%) or a small change (0.54\%) for most of cases, except for an obvious increase under the 100\% profile discard ratio.
We think that the reason is that some of the predicted profiles are not even in provided profiles, which leads to a very limited \acs{PEA} score of 34.78\% under the 100\% profile discard ratio (see Table~\ref{tab:analysis_ablation_study_UPE}). 
But these predicted values happen to be useful for selecting an appropriate response in \acs{DRS}.

\subsection{Effect of multiple-hop mechanism (Q2)}
We compare the \acs{RSA} performance of \acs{CoMemNN} and \acs{NPMemNN} with different numbers of hops.
The results are shown in Table~\ref{tab:analysis_hop}. 
\vspace{-0.04\linewidth}
\begin{table}[htb!]
\setlength{\tabcolsep}{13.2pt}
\centering
\caption{Analysis of the effect of hop number on \acs{DRS}. \textbf{Bold face} indicates leading results. Significant improvements over \acs{NPMemNN} are marked with $^\ast$ (paired t-test, $p < 0.01$). The values of Diff. are computed by absolute difference of \acs{RSA} (\%) between \acs{CoMemNN} and \acs{NPMemNN}.}
\label{tab:analysis_hop}
\begin{tabular}{@{}ccccc@{}}
\toprule
\textbf{\#Hop} & \bf 1 & \bf 2 & \bf 3 & \bf 4 \\ 
\midrule
\acs{NPMemNN} & 88.11 & 87.22 & 87.91   & 87.61 \\
\acs{CoMemNN} & \textbf{90.07}\rlap{$^\ast$} & \textbf{90.78}\rlap{$^\ast$} & \textbf{91.13}\rlap{$^\ast$} & \textbf{90.77}\rlap{$^\ast$}  \\ 
\midrule
Diff. & \phantom{0}1.96 & \phantom{0}3.56 & \phantom{0}3.22 & \phantom{0}3.16\\
% Large & 98.13 & 98.16 & xxx & xxx \\
\bottomrule
\end{tabular}
\end{table}

We see that \acs{CoMemNN} greatly outperforms \acs{NPMemNN} by a large margin (1.96\%--3.56\%) with all number of hops.
This further confirms the non-trivial improvement of  \acs{CoMemNN}.
Besides, \acs{CoMemNN} improves by 1.06\% when the number of hop changes from 1 to 3 and slightly decreases with 4.
This means that \acs{CoMemNN} benefits from a multiple-hop mechanism.

\subsection{Effect of different profile attributes (Q3)}
We explore how the four types of profile attributes (i.e., gender, age, dietary preference, and favorite food) affect the \acs{RSA} performance.
The results are shown in Table~\ref{tab:analysis_attribute_importance_1}.
\begin{table}[htb!]
\setlength{\tabcolsep}{2pt}
\centering
\caption{Analysis of profile attribute importance to \acs{DRS}. 
\textbf{Discard attribute} table shows we discard all values of a specific attribute or a combination of two specific attributes. 
\textbf{Retain attribute} table shows we retain all values of a specific attribute and discard all values for the rest.
Underline indicates the lower bound baseline that retains no attributes.
Bold face indicates the upper bound baseline that retains all attributes.}
\label{tab:analysis_attribute_importance_1}
\begin{tabular}{@{}lcccccc@{}}
\toprule
\textbf{Discarded attribute} & none  & gender & age   & dietary & favorite & all   \\ \midrule
gender                     & /     & \textit{93.05}  & \textit{91.94} & 88.86   & \textit{91.95}    & /     \\
age                        & /     & /      & \textit{92.26} & 89.37   & 91.04    & /     \\
dietary                    & /     & /      & /     & 86.74   & 86.42    & /     \\
favorite                   & /     & /      & /     & /       & 90.25    & /     \\ 
\midrule
\textbf{Retained attribute}  & \underline{82.85} & 87.46  & 87.93 & 90.57   & 87.37    &
\textbf{91.13} \\ 
\bottomrule
\end{tabular}
\end{table}

First, each attribute works well in isolation.
Specifically, when we only retain the values of each single attribute, we obtain the results in the last row as 87.46\%, 87.93\%, 90.57\%, 87.37\% for gender, age, dietary, favorite, respectively. 
% We can see that all four types of attributes improve the \acs{RSA} performance separately and the \acs{RSA} score is between the lower bound (82.25\%) and the upper bound (91.13\%).
The attribute ``dietary'' contributes most followed by ``age'', ``gender'' and ``favorite.''

Second, different types of attributes depend on each other and influence the \acs{RSA} performance differently.
If we only remove the values of one attribute, we get the results on the diagonal: 93.05\%, 92.26\%, 86.74\%, 90.25\%, respectively.
Removing ``dietary'' drops most followed by ``favorite.''
Thus, ``dietary'' contributes more than the rest.

An exception is that the \acs{RSA} performance increases when discarding ``gender'' and ``age.''
We believe this is the effect of the neighbors.
To show this, we further investigate the effect of ``gender'' and ``age'' without using neighbor information.
The results are shown in Table~\ref{tab:analysis_attribute_importance_2}.
\begin{table}[htb!]
\setlength{\tabcolsep}{22pt}
\centering
\caption{Analysis of profile attribute importance to \acs{DRS} without the effect of neighbors. 
\textbf{Bold face} indicates the baseline of \acs{CoMemNN} without neighbors.
In each cell, the ﬁrst number represents the \acs{RSA} (\%), 
and the number in parenthesis shows the difference values, and $\downarrow$ denotes decrease compared with the baseline.}
\label{tab:analysis_attribute_importance_2}
\begin{tabular}{@{}lc@{}}
\toprule
                                             & \acs{RSA} (Diff.)\\ \midrule
\acs{CoMemNN} w/o neighbors                & \textbf{90.34}                                    \\ \midrule
\acs{CoMemNN} w/o neighbors - gender       & 88.25 ($\downarrow$2.09)                                 \\
\acs{CoMemNN} w/o neighbors - age          & 85.62 ($\downarrow$4.72)                                 \\
\acs{CoMemNN} w/o neighbors - gender - age & 83.73 ($\downarrow$6.61) \\ \bottomrule
\end{tabular}
\end{table}
%

% We remove all neighbors from \acs{CoMemNN} and get the new upper bound baseline as 90.34\%. 
We can see that removing ``gender'' and ``age'' decreases the performance in this case.
Thus, the different effects of ``gender'' and ``age'' are due to the neighbors.

% To sum up, different type of attributes affect \acs{RSA} performance differently in both direct and indirect ways.
% In the former, this purely affects on the process of selecting a response, specially in this work, dietary contributes most followed by age, gender and favorite.
% In the later, this also impacts the quality of the retrieved neighbors.

%!TEX root = ./WWW21-Jiahuan-main.tex

\section{Conclusion}
In this paper, we have studied personalized \acp{TDS} without assuming that we have complete user profiles. 
We have proposed \acf{CoMemNN}, which introduces a cooperative mechanism to enrich user profiles gradually as dialogues progress, and to improve response selection based on enriched profiles simultaneously.
We also devise a learning algorithm to effectively learn \acs{CoMemNN} with multiple hops.

Extensive experiments on the \acf{PbAbI} dataset demonstrate that \acs{CoMemNN} significantly outperforms state-of-the-art baselines.
Further analysis experiments confirm the effectiveness of \acs{CoMemNN} by analyzing the performance and contribution of each component.

A limitation of our work is that we tested the performance of \acs{CoMemNN} on the only open available personalized \acp{TDS} dataset \ac{PbAbI}. 
We encourage the community to work on creating additional resources for this task.

As to future work, we hope to experiment on more datasets and investigate how the performance varies on different datasets and whether we can further improve the performance by leveraging non-personalized \ac{TDS} datasets.

\begin{acks}
This research was partially supported by
the China Scholarship Council (CSC).
All content represents the opinion of the authors, which is not necessarily shared or endorsed by their respective employers and/or sponsors.
\end{acks}

% \clearpage
\bibliographystyle{ACM-Reference-Format}
\bibliography{references}

% \appendix

\end{document}